\documentclass{bmvc2k}
\usepackage{amsfonts,amssymb,amsmath,amsthm,bbm,pifont,ulem}
\usepackage{enumitem}
\usepackage{wrapfig}
\usepackage{hyperref}

\usepackage{algorithm,algpseudocode,amsmath}

\title{Polycentric Clustering and Structural Regularization for Source-free\\ Unsupervised Domain Adaptation}

\addauthor{Xinyu Guan}{xinyuu@nuaa.edu.cn}{1}
\addauthor{Han Sun}{sunhan@nuaa.edu.cn}{1}
\addauthor{Ningzhong Liu}{liunz@163.com}{1}
\addauthor{Huiyu Zhou}{hz143@leicester.ac.uk}{2}

\addinstitution{
 Nanjing University of Aeronautics and Astronautics\\
 Nanjing, China
}
\addinstitution{
 University of Leicester\\
 UK
}

\runninghead{Student, Prof, Collaborator}{BMVC Author Guidelines}

\begin{document}

\maketitle

\begin{abstract}
Source-Free Domain Adaptation (SFDA) aims to solve the domain adaptation problem by transferring the knowledge learned from a pre-trained source model to an unseen target domain. Most existing methods assign pseudo-labels to the target data by generating feature prototypes. However, due to the discrepancy in the data distribution between the source domain and the target domain and category imbalance in the target domain, there are severe class biases in the generated feature prototypes and noisy pseudo-labels. Besides, the data structure of the target domain is often ignored, which is crucial for clustering. In this paper, a novel framework named PCSR is proposed to tackle SFDA via a novel intra-class {\bf P}olycentric {\bf C}lustering and {\bf S}tructural {\bf R}egularization strategy. Firstly, an inter-class balanced sampling strategy is proposed to generate representative feature prototypes for each class. Furthermore, k-means clustering is introduced to generate multiple clustering centers for each class in the target domain to obtain robust pseudo-labels. Finally, to enhance the model's generalization, structural regularization is introduced for the target domain. Extensive experiments on three UDA benchmark datasets show that our method performs better or similarly against the other state of the art methods, demonstrating our approach's superiority for visual domain adaptation problems.
\end{abstract}

\section{Introduction}
\label{sec:intro}
In recent years, unsupervised domain adaptation~\cite{ganin2015unsupervised} has been developed to reduce the domain shift by transferring knowledge from a labeled source domain to a target domain, and has achieved promising results in object detection~\cite{chen2018domain,su2020adapting}, semantic segmentation~\cite{toldo2020unsupervised,zou2018unsupervised} and person re-identification~\cite{deng2018image,wang2019beyond}. The main research directions of the existing UDA methods include (i) minimizing the distribution discrepancy by matching the statistical distribution moments~\cite{long2018transferable,peng2019moment}; (ii) applying adversarial training to learn domain-invariant feature representations~\cite{xu2020reliable,zhang2019domain}; and (iii) bridging the domain gap by using clustering~\cite{deng2019cluster} or pseudo-labeling~\cite{arazo2020pseudo}. It is worth noting that all of them assume that both well-trained source models and labeled source data are available. However, with the increasing concerns about data privacy and intellectual property of users, the accessibility of well-labeled source data cannot be guaranteed for many real-world tasks.

To overcome the above problem, some recent works~\cite{li2020model,kundu2020universal,liang2020we,kim2021domain,sahoo2020unsupervised,yang2020unsupervised} explored domain adaptation without source data. Source-free domain adaptation (SFDA) is a new unsupervised learning setup for the domain adaptation task. Recently, image generation~\cite{li2020model}, class prototypes~\cite{liang2020we,yang2020unsupervised}, and pseudo labeling~\cite{liang2020we} are widely utilized in the existing SFDA approaches. However, generative models require a large computational capacity for generating target-style images. Class prototypes and pseudo labeling methods have shown competitive results but noisy labels are introduced due to category biases in the source and target domains. We argue that only one clustering center for each category is insufficient to avoid negative transfer caused by hard transfer data in the target domain. Furthermore, the structural information of the target data in the feature space is often ignored, which is very helpful to reduce the noisy labels though.

Based on the ideas presented above, a simple yet effective structured clustering strategy for polycentric clustering is proposed in this paper. Specifically, due to class imbalance in the target domain, to avoid the easy data dominating the target model, an inter-class-balanced sampling strategy is designed to aggregate representative samples of each class. To assign more accurate pseudo-labels, polycentric clustering is proposed to generate multiple feature clustering centers within a class of the target data. In addition, to alleviate the noisy labels, a mixup structural regularization term is introduced into our framework, encouraging the interpolation samples to be consistent with the interpolation predictions. Under the guidance of structural regularization, the model is enforced to maintain consistency, thus the robustness against noisy labels is improved. 

To evaluate the effectiveness of our model, we conduct extensive experiments on three benchmark datasets, and the experimental results show significant superiority of our method in SFDA. The main contributions of our work are summarized as follows:
\setenumerate[1]{itemsep=0pt,partopsep=0pt,parsep=\parskip,topsep=0pt}
\setitemize[1]{itemsep=0pt,partopsep=0pt,parsep=\parskip,topsep=0pt}
\setdescription{itemsep=0pt,partopsep=0pt,parsep=\parskip,topsep=0pt}
\begin{itemize}
\item We propose a novel framework, Polycentric Clustering and Structure Regularization (PCSR) for SFDA tasks, which aims to protect data privacy and maintain the model performance without access to the source data.
\item To avoid easy-transfer data dominating the target model, an inter-class-balanced sampling strategy is designed to address the challenge of class imbalance. And a polycentric clustering approach is proposed for each class to reduce the noisy labels for those hard data.
\item To reduce the noisy labels, the mixup regularization module is introduced to interpolate the target data for consistent training, leading to more robust pseudo labels.
\item Extensive experiments on three benchmark datasets validate the superiority of our PCSR strategy. The results show that our strategy is comparable to or significantly outperform existing methods.
\end{itemize}

\section{Related Work}
\subsection{Unsupervised Domain Adaptation}
As a classic example of transfer learning~\cite{pan2009survey}, in recent years, UDA methods for image classification tasks have aimed to align the source and target distributions in an attempt to minimize the domain gap in knowledge transfer. There are currently three main classes of UDA methods: discrepancy-, adversarial learning-, and reconstruction-based. The feature distributions of the source and target domains are aligned by minimizing Maximum Mean Discrepancy (MMD)~\cite{long2015learning,long2016unsupervised,long2017deep} in the discrepancy-based methods. In the adversarial learning-based approaches, the network is trained to learn domain invariant features by adding a feature discriminator~\cite{ganin2016domain,long2018conditional,shu2018dirt}. Different from the previous two methods, the network is guided to extract domain-invariant features by an auxiliary image reconstruction module in the reconstruction-based approaches~\cite{murez2018image,bousmalis2016domain}. Although these UDA methods are effective, they require access to both the source and target data. In the real world, this is impractical due to data privacy or security concerns. In contrast, our method proposed in this paper does not require source data when performing adaptation, making it more suitable for real-world applications.

\subsection{Unsupervised Source-Free Domain Adaptation}
In most realistic scenarios, only source models and unlabeled target data are available. As a result, some recent work on source-free domain adaptation has emerged~\cite{li2020model,kundu2020universal,liang2020we,yang2021generalized,2021Source,xia2021adaptive,liu2021source,tian2021vdm}. Specifically, SHOT~\cite{liang2020we} proposed freezing the source classifier to maximize mutual information and minimize entropy, while using a pseudo-labeling strategy to obtain extra supervision. In 3C-GAN~\cite{li2020model}, labeled target-style training images were generated based on a conditional GAN to improve model performance on the target domain. In G-SFDA~\cite{yang2021generalized}, the neighborhood structure of the target data for clustering enhanced the predictive consistency of local neighborhood features effectively. In CPGA~\cite{2021Source}, the source avatar prototypes were generated via contrastive learning to mine the hidden knowledge in the source model. Many methods described above freeze the source classifier during adaptation to preserve class information and assign pseudo-labels based on the classifier’s output. They mainly focus on a single feature prototype to align two domains, which often causes negative transfer and noisy labels. Instead, we here introduce polycentric clustering for each class to reduce noisy labels. In addition, a consistent training strategy is introduced to enhance the target domain for source-free domain adaptation.

\section{Method}
In this section, we first formally define the problem and the notation used for source-free domain adaptation followed by an overview of our framework. Later, a detailed description of our proposed strategy to solve the SFDA problem is presented.

\subsection{Preliminaries and notations}
We denote that the source domain with $n_s$ labeled samples as $\mathcal{D}_s=\{{x_s^i},{y_s^i}\}_{i=1}^{n_s}$, where ${x_s^i\in{\mathcal{X}_s}}$, ${y_s^i\in{\mathcal{Y}_s}\subseteq{\mathbb{R}^K}}$ is the one-hot ground-truth label and K is the total number of classes of label set $\mathcal{C}=\{1,2,...,K\}$. Target domain dataset $\mathcal{D}_t$ has $n_t$ unlabeled samples $\{{x_t^i}\}_{i=1}^{n_t}$, where ${x_t^i\in{\mathcal{X}_t}}$, it has the same label set $\mathcal{C}$ as that of $\mathcal{D}_s$. Under the SFDA setting, only the model $f_s$ trained on the source data is accessible, which consists of two parts: a feature extractor $g_s\colon{\mathcal{X}_s}\rightarrow{\mathbb{R}^d}$ and a classifier $h_s\colon{\mathbb{R}^d}\rightarrow{\mathbb{R}^K}$, i.e., $f_s(x)=h_s(g_s(x))$, where $d$ denotes the dimension of the feature space. In this work, with only the source model $f_s$ and unlabeled data $\{{x_t^i}\}_{i=1}^{n_t}$ available, our goal is to learn an objective function $f_t\colon{\mathcal{X}_t}\rightarrow{\mathcal{Y}_t}$ and to infer $\{{y_t^i}\}_{i=1}^{n_t}$.

\subsection{Overall framework}
\begin{figure}
	\centering
	\includegraphics[width=1.0\textwidth]{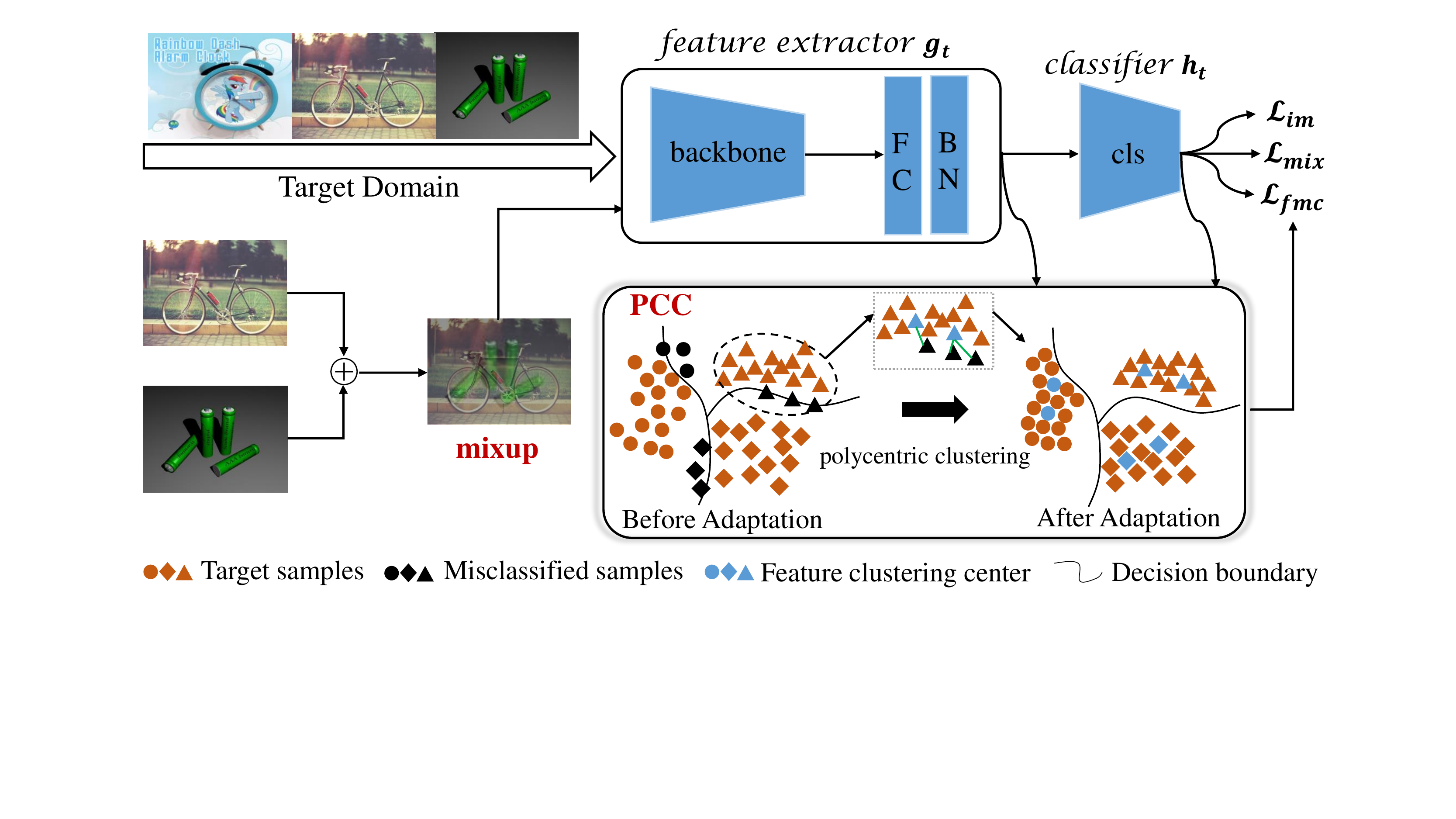}\\[-1pt]
	\setlength{\belowcaptionskip}{-0.4cm} 
	\setlength{\abovecaptionskip}{-2.2cm}
	\caption{Overall training pipeline of our proposed PCSR method.}
	\label{fig:fig1}
\end{figure}

An overview of our proposed framework is presented in Figure \ref{fig:fig1}. The target model $f_t$ is initialized by the source model $f_s$, and the source model consists of two modules: the feature encoding module $g_s$ and the classifier module $h_s$. The target model uses the same classifier module, namely, $h_t=h_s$, and two new modules named PCC and mixup are introduced respectively. It is noted that our PCSR learn $f_t$ in an epoch-wise manner. As for each epoch stage, firstly, a balanced set of feature instances representing each class is obtained using inter-class balanced sampling. Then polycentric clustering is implemented to obtain accurate pseudo-labeling. After that, information maximization loss is used to reduce the gap between the feature distributions in the source and target domains. Meanwhile, the mixup operation is introduced to enhance the target domain with more interpolated samples.

\subsection{Information maximization}
We update the feature extractor $g_t$ using the information maximization (IM) loss~\cite{hu2017learning}, which reduces the feature distribution between the source and target domains so that the classification output of the target features has some certainty and global diversity. The IM loss consists of a conditional entropy term and a diversity term:
\begin{equation}
\label{eq1}
\mathcal{L}_{im}=-\mathbb{E}_{x\in{\mathcal{X}_t}}\sum_{k=1}^K\delta_k(f_t(x))\log\delta_k(f_t(x))+\sum_{k=1}^K\bar{p}_k\log\bar{p}_k
\end{equation}
Where $\delta_k(a)$ denotes the $k$-th element in the softmax output of the K-dimensional vector $a$. $\bar{p}=\mathbb{E}_{x\in{\mathcal{X}_t}}[\delta_k(f_t(x))]$ is the average of the current batch’s softmax output. 
\subsection{Intra-class polycentric clustering}
To reduce the gap between the source and target domains, a simple existing solution is to eliminate the noisy labels by selecting pseudo-labels with high confidence. However, this will bias the model toward majority classes and ignore minority classes, resulting in noisy labels for those hard data in the target domain. To reduce the noisy labels, an intra-class polycentric clustering strategy is proposed, which contains two steps.

{\bf Inter-class balanced sampling}. Due to class imbalance in the target domain, instead of using the existing prediction results based on argmax operations, we adopt an inter-class balanced sampling strategy to construct each class of the target domain. Specifically, for the $k$-th class in the target domain, each sample in the target domain is represented by a feature vector $\hat{g}_t(x_t)$ and a classification result $p(x_t)=\delta(\hat{f}_t(x_t))$. Instead of choosing the top-1 feature, the top-M $p(x_t)$ of the $k$-th class on the target domain $\mathcal{D}_t$ are selected as potential representative features for aggregation. Then these top-M are averaged to form an inter-class balanced feature clustering center $c_k$, and the initial pseudo-label $\hat{y}_t$ is obtained from the nearest centroid classifier as follows:
\begin{equation}
\label{eq2}
\begin{aligned}
\mathcal{M}_k&=\mathop{\arg\max}\limits_{x\in{\mathcal{X}_t}}\delta_k(\hat{f}_t(x)) \\
c_k^{(0)}&=\frac{1}{M}\sum_{i\in{\mathcal{M}_k}}\hat{g}_t(x_i) \\ 
\hat{y}_t&=\mathop{\arg\min}\limits_{k}D_f(\hat{g}_t(x),c_k^{(0)})
\end{aligned}
\end{equation}
Where $\hat{f}_t=\hat{g}_t\circ{h_t}$ denotes the previously learned target hypothesis, $M=\max{(1,\lfloor\frac{n_t}{r\times{K}}\rfloor)}$. $r$ is the hyperparameter of the top-M selection ratio and K is the number of classes in the target domain. $D_f(a,b)$ measures the cosine distance between $a$ and $b$. Based on the above strategy, we can obtain balanced sampled feature instances for each class. Similar to SHOT~\cite{liang2020we}, we perform iterative computations to obtain more robust clustering centers $c_k$ and pseudo-labels $\hat{y}_t$ as follows:
\begin{equation}
\label{eq3}
\begin{aligned}
\mathcal{M}_k&=\mathop{\arg\max}\limits_{x\in{\mathcal{X}_t}}\delta_k(\hat{g}_t(x)\cdot{c_k^{(0)}}) \\
c_k^{(1)}&=\frac{1}{M}\sum_{i\in{\mathcal{M}_k}}\hat{g}_t(x_i) \\ 
\hat{y}_t&=\mathop{\arg\min}\limits_{k}D_f(\hat{g}_t(x),c_k^{(1)})
\end{aligned}
\end{equation}
Although the pseudo-labels and the centroids can be updated by Eq. (\ref{eq3}) multiple times, we find that two rounds of updating are sufficient.

{\bf Polycentric clustering}. According to the above strategy, class-balanced prototype and robust pseudo-labels can be obtained. However, for those ambiguous data located nearby the decision boundary, they may not be effectively represented by a coarse monocentric prototype. In this paper, polycentric clustering is proposed to get more accurate pseudo-labels with a predefined number of clustering centers. Specifically, the classical k-means algorithm~\cite{macqueen1967classification} is introduced to achieve intra-class clustering of the target domain, assuming that the number of clustering centers is P and $\{c_k^i\}_{i=1}^P$ is defined as the multiple clustering centers of the $k$-th class. The k-means algorithm is used to obtain multiple clustering centers of each class $\{c_k^i\}_{i=1}^P$ and obtain more robust pseudo-labels $\hat{y}_t$:
\begin{equation}
\label{eq4}
\begin{aligned}
\mathcal{M}_k &= \mathop{\arg\max}\limits_{x\in{\mathcal{X}_t}} \frac{\mathop{\max}\limits_{1\leqslant i\leqslant P}(\exp(\hat{g}_t(x)\cdot{c_k^i}))}
{\sum_{j=1}^K \mathop{\max}\limits_{1\leqslant i\leqslant P}(\exp(\hat{g}_t(x)\cdot{c_j^i}))} \\
\{c_k^i\}_{i=1}^P &= \mathop{Kmeans}\limits_{m\in{\mathcal{M}_k}}(\hat{g}_t{(x_m)}) \\
\hat{y}_t &= \mathop{\arg\max}\limits_{k} \frac{\mathop{\max}\limits_{1\leqslant i\leqslant P}(\exp(\hat{g}_t(x)\cdot{c_k^i}))}
{\sum_{j=1}^K \mathop{\max}\limits_{1\leqslant i\leqslant P}(\exp(\hat{g}_t(x)\cdot{c_j^i}))}
\end{aligned}
\end{equation}
Empirically, we find that iterating this process for two rounds is sufficient. Given the generated pseudo-labeling, the loss function for computing the intra-class polycentric clustering pseudo-labeling is as follows.
\begin{equation}
\label{eq5}
\mathcal{L}_{pcc}=-\mathbb{E}_{x\in{\mathcal{X}_t}}\sum_{k=1}^K\mathbbm{1}_{[\hat{y}_t=k]}\log\delta_k(f_t(x))
\end{equation}

\subsection{Structural regularization by mixup training}
As mentioned above, we consider intra-class polycentric clustering to mitigate negative transfer, but this ignores the target domain’s data structure and still suffers from the noisy labels. According to~\cite{yang2021exploiting}, even though the target data is shifted in the feature space, the target data of the same class is still expected to form a cluster in the embedding space. Therefore, we consider paired target structure information by MixUp~\cite{zhang2017mixup} to reduce the intra-domain variation, and the new instance $\{x,y\}$ generated by the MixUp operation $Mix((X_1,Y_1),(X_2,Y_2))$ can be defined as:
\begin{equation}
\label{eq6}
\begin{aligned}
x &= \lambda{x_1}+(1-\lambda)x_2;\quad x_1\in{X_1},\;x_2\in{X_2} \\
y &= \lambda{y_1}+(1-\lambda)y_2;\quad y_1\in{Y_1},\;y_2\in{Y_2}
\end{aligned}
\end{equation}
$\lambda$ denotes the mixup coefficient. The structured loss is obtimized by using interpolation consistency training~\cite{verma2019interpolation}:
\begin{equation}
\label{eq7}
\mathcal{L}_{mix}=\mathbb{E}_{x_i,x_j\in{\mathcal{X}_t}}l_{ce}((\lambda f_t^{'}(x_i)+(1-\lambda)f_t^{'}(x_j)),f_t(\lambda x_i)+(1-\lambda)x_j)
\end{equation}
Where $\lambda$ obeys Beta distribution sampling, $\lambda\in{Beta(\alpha,\alpha)}$, and the hyperparameter is empirically set to 0.3, following the setup of~\cite{zhang2017mixup}. $l_{ce}$ represents the cross-entropy loss. $f_t^{'}$ indicates that no gradient calculation is required, but only the value of ${f_t}$ is provided. This loss function can supply more augmented samples for the target domain, allowing for better generalization ability.

Integrating all the loss function equations introduced, we can derive the final loss function as follows.
\begin{equation}
\label{eq8}
\mathcal{L}_{t}=\mathcal{L}_{im}+\mathcal{L}_{pcc}+\beta\mathcal{L}_{mix}
\end{equation}
Where $\beta$ is a hyperparameter experimentally set to 1.0. 

Algorithm \ref{alg1} summarizes our method's training process.
\begin{algorithm}
  \caption{{\bf P}olycentric {\bf C}lustering and {\bf S}tructural {\bf R}egularization for SFDA} 
  \label{alg1}
  \begin{algorithmic}[1]
  \Require  
      Pre-trained source model $f_s(h_s,g_s)$;  
      target data $\mathcal{X}_t$;
      max epoch number $T_{max}$;
      Number of clustering centers P. 
    \Ensure  
      Initialize $f_t(h_t,g_t)$ using $f_s(h_s,g_s)$. 
    \For{$epoch\_idx=1$ to $T_{max}$}  
      \State Obtain feature prototypes and initial
      pseudo-labels using inter-class balanced sampling using Eq. (\ref{eq2}) ~ Eq.(\ref{eq3}).
      \State Compute multiple clustering centers and pseudo-labels based on P and k-means algorithm using Eq.(\ref{eq4}).
      \For{$iter\_idx=1$ to the number of target samples $N_b$}  
        \State Calculate IM loss according to the Eq.(\ref{eq1}).
        \State Apply MixUp to performing structural regularization operations by Eq.(\ref{eq7}).
      \EndFor
      \State Update $f_t$ via minimizing Eq.(\ref{eq8}).
    \EndFor  
  \end{algorithmic}  
\end{algorithm} 

\section{Experiment and Analysis}
\subsection{Experimental setup}
{\bf DataSets}. We conduct experiments on three datasets, including Office-31~\cite{saenko2010adapting}, Office-Home~\cite{venkateswara2017deep}, and VisDA-C~\cite{peng2017visda}. {\bf Office-31} is divided into three domains: Amazon(A), Webcam(W), and DSLR(D), with 31 categories. {\bf Office-Home} contains 65 categories and consists of four domains: Artistic images(A), Clip Art(C), Product images(P), and Real-World images(R). {\bf VisDA-C} is a more challenging dataset, with 152K synthetic images generated by rendering 3D models in the source domain while the target domain has 55K real object images, which are divided into 12 shared classes.

{\bf Implementation details}. To ensure a fair comparison with the related approaches, we employ ResNet-50~\cite{he2016deep} pre-trained on Image-Net~\cite{deng2009imagenet} as the backbone for Office-31 and Office-Home, and ResNet-101~\cite{he2016deep} as the backbone for VisDA-C. Similar to the previous work, for all datasets, we apply the gradient descent (SGD) optimizer with momentum 0.9 and weight decay 1e-3, the batch size is set to 64, and the input image size is reshaped to 224×224. The learning rate is set to 1e-2 for Office-31 and Office-Home, and 1e-3 for VisDA-C, and 30 epochs are trained for all the settings. For the hyperparameter settings, we set the hyperparameter $r$ of the selection ratio to 3 on all datasets, and the predefined number of clustering centers P is set to 3 for Office-31 and Office-Home, and 4 for VisDA-C. All experiments are built on a TITAN Xp with Pytorch-3.8. The source code of the proposed algorithm 
is available in  \href{https://github.com/Gxinuu/PCSR}{https://github.com/Gxinuu/PCSR}.

\subsection{Quantitative comparison}
Tables 1-3 show the experimental results on the three datasets mentioned above. The best results in SFDA shown in {\bf bolded} font and the sub-optimal results \uline{underlined}. In Table \ref{tab:tab1}, our method achieves comparable results to 3C-GAN on Office-31 and even obtains more competitive performance. Note that 3C-GAN highly relies on the extra synthesized data. And Office-31 is a small-scale dataset whose image number of each class is around 40 on average. Therefore, it is hard for our method to aggregate valid polycentric clustering. Yet we still achieve the best results on 3 of 6 tasks.
\begin{table}
\begin{center}
\renewcommand\arraystretch{0.9}
\resizebox{.7\textwidth}{!}{
\begin{tabular}{lcccccccc}
    \hline
    \multicolumn{1}{c}{Method} & 
    SF & 
    \multicolumn{1}{c}{A→D} & 
    \multicolumn{1}{c}{A→W} & 
    \multicolumn{1}{c}{D→A} & 
    \multicolumn{1}{c}{D→W} &
    \multicolumn{1}{c}{W→A} & 
    \multicolumn{1}{c}{W→D} & 
    \multicolumn{1}{c}{Avg.} \\
    \hline
    CDAN(2018b)\cite{long2018conditional} & \ding{55} & 92.9  & 94.1  & 71.0  & 98.6  & 69.3  & 100.0  & 87.7  \\
    SAFN(2019)\cite{xu2019larger} & \ding{55} & 88.8  & 98.4  & 69.8  & 99.8  & 69.7  & 87.7  & 85.7  \\
    MCC(2020)\cite{jin2020minimum} & \ding{55} & 95.6  & 95.4  & 72.6  & 98.6  & 73.9  & 100.0  & 89.4  \\
    BNM(2020)\cite{cui2020towards} & \ding{55} & 90.3  & 91.5  & 70.9  & 98.5  & 71.6  & 100.0  & 87.1  \\
    SRDC(2020)\cite{tang2020unsupervised} & \ding{55} & 95.8  & 95.7  & 76.7  & 99.2  & 77.1  & 100.0  & 90.8  \\
    DMRL(2020)\cite{wu2020dual} & \ding{55} & 93.4  & 90.8  & 73.0  & 99.0  & 71.2  & 100.0  & 87.9  \\
    RWOT(2020)\cite{xu2020reliable} & \ding{55} & 94.5  & 95.1  & 77.5  & 99.5  & 77.9  & 100.0  & 90.8  \\
    LAMDA(2021)\cite{le2021lamda} & \ding{55} & 96.0  & 95.2  & 87.3  & 98.5  & 84.4  & 100.0  & 93.6  \\
    \hline
    Source-only & \checkmark & 80.7  & 77.0  & 60.8  & 95.1  & 62.3  & 98.2  & 79.0  \\
    SFDA(2021)\cite{kim2021domain} & \checkmark & 92.2  & 91.1  & 71.0  & 98.2  & 71.2  & 99.5  & 87.2  \\
    SHOT(2020a)\cite{liang2020we} & \checkmark & \uline{94.0}  & 90.1  & 74.7  & 98.4  & 74.3  & \uline{99.9}  & 88.6  \\
    BAIT(2020)\cite{yang2020unsupervised} & \checkmark & 92.0  & \textbf{94.6} & 74.6  & 98.1  & \uline{75.2}  & \textbf{100.0} & 89.1  \\
    3C-GAN(2020)\cite{li2020model} & \checkmark & 92.7  & 93.7  & 75.3  & \uline{98.5}  & \textbf{77.8} & 99.8  & \textbf{89.6} \\
    NRC(2021a)\cite{yang2021exploiting} & \checkmark & \textbf{96.0} & 90.8  & 75.3  & \textbf{99.0} & 75.0  & \textbf{100.0} & 89.4  \\
    BNM-S(2021)\cite{cui2021fast} & \checkmark & 93.0  & 92.9  & \uline{75.4}  & 98.2  & 75.0  & \uline{99.9}  & 89.1  \\
    \bf{ours} & \checkmark & 93.6  & \uline{93.8}  & \textbf{76.0} & \textbf{99.0} & 74.5  & \textbf{100.0} & \uline{89.5}  \\
    \hline
\end{tabular}%
}
\end{center}
\caption{Classification accuracies (\%) on Office-31 for ResNet50-based methods.}
\label{tab:tab1}
\end{table}%

\begin{table}
\begin{center}
\setlength{\tabcolsep}{1pt}
\renewcommand\arraystretch{0.9}
\resizebox{\linewidth}{!}{
\begin{tabular}{lcccccccccccccc}
    \hline
    \multicolumn{1}{c}{Method} & 
    SF & 
    \multicolumn{1}{c}{Ar→Cl} & 
    \multicolumn{1}{c}{Ar→Pr} &
    \multicolumn{1}{c}{Ar→Re} & 
    \multicolumn{1}{c}{Cl→Ar} & 
    \multicolumn{1}{c}{Cl→Pr} & 
    \multicolumn{1}{c}{Cl→Re} & 
    \multicolumn{1}{c}{Pr→Ar} & 
    \multicolumn{1}{c}{Pr→Cl} & 
    \multicolumn{1}{c}{Pr→Re} & 
    \multicolumn{1}{c}{Re→Ar} & 
    \multicolumn{1}{c}{Re→Cl} & 
    \multicolumn{1}{c}{Re→Pr} & 
    \multicolumn{1}{c}{Avg.} \\
    \hline
    ResNet-50(2016)\cite{he2016deep} & \ding{55} & 34.9  & 50.0  & 58.0  & 37.4  & 41.9  & 46.2  & 38.5  & 31.2  & 60.4  & 53.9  & 41.2  & 59.9  & 46.1  \\
    CDAN(2018b)\cite{long2018conditional} & \ding{55} & 50.7  & 70.6  & 76.0  & 57.6  & 70.0  & 70.0  & 57.4  & 50.9  & 77.3  & 70.9  & 56.7  & 81.6  & 65.8  \\
    BNM(2020)\cite{cui2020towards} & \ding{55} & 52.3  & 73.9  & 80.0  & 63.3  & 72.9  & 74.9  & 61.7  & 49.5  & 79.7  & 70.5  & 53.6  & 82.2  & 67.9  \\
    SAFN(2019)\cite{xu2019larger} & \ding{55} & 52.0  & 71.7  & 76.3  & 64.2  & 69.9  & 71.9  & 63.7  & 51.4  & 77.1  & 70.9  & 57.1  & 81.5  & 67.3  \\
    SRDC(2020)\cite{tang2020unsupervised} & \ding{55} & 52.3  & 76.3  & 81.0  & 69.5  & 76.2  & 78.0  & 68.7  & 53.8  & 81.7  & 76.3  & 57.1  & 85.0  & 71.3  \\
    LAMDA(2021)\cite{le2021lamda} & \ding{55} & 57.2  & 78.4  & 82.6  & 66.1  & 80.2  & 81.2  & 65.6  & 55.1  & 82.8  & 71.6  & 59.2  & 83.9  & 72.0  \\
    \hline
    Source-only & \checkmark & 44.0  & 67.0  & 73.5  & 50.7  & 60.3  & 63.6  & 52.6  & 40.4  & 73.5  & 65.7  & 46.2  & 78.2  & 59.6  \\
    SFDA(2021)\cite{kim2021domain} & \checkmark & 48.4  & 73.4  & 76.9  & 64.3  & 69.8  & 71.7  & 62.7  & 45.3  & 76.6  & 69.8  & 50.5  & 79.0  & 65.7  \\
    SHOT(2020a)\cite{liang2020we} & \checkmark & 57.1  & 78.1  & 81.5  & \uline{68.0}  & 78.2  & 78.1  & 67.4  & 54.9  & 82.2  & 73.3  & 58.8  & 84.3  & 71.8  \\
    BAIT(2020)\cite{yang2020unsupervised} & \checkmark & 57.4  & 77.5  & \textbf{82.4} & \uline{68.0}  & 77.2  & 75.1  & 67.1  & 55.5  & 81.9  & \uline{73.9}  & 59.5  & 84.2  & 71.6  \\
    G-SFDA(2021b)\cite{yang2021generalized} & \checkmark & 57.9  & 78.6  & 81.0  & 66.7  & 77.2  & 77.2  & 65.6  & 56.0  & 82.2  & 72.0  & 57.8  & 83.4  & 71.3  \\
    NRC(2021a)\cite{yang2021exploiting} & \checkmark & 57.7  & \textbf{80.3} & 82.0  & \textbf{68.1} & \textbf{79.8} & 78.6  & 65.3  & \uline{56.4}  & \textbf{83.0} & 71.0  & 58.7  & \textbf{85.6} & \uline{72.2}  \\
    BNM-S(2021)\cite{cui2021fast} & \checkmark & 57.4  & 77.8  & 81.7  & 67.8  & 77.6  & \textbf{79.3} & \uline{67.6}  & 55.7  & 82.2  & 73.5  & 59.5  & \uline{84.7}  & 72.1  \\
    \bf{ours} & \checkmark & \textbf{58.1} & 78.5  & \uline{82.1}  & 67.9  & \uline{79.1}  & \uline{78.8}  & \textbf{69.0} & \textbf{57.9} & \uline{82.3}  & \textbf{75.2} & \textbf{60.0} & \uline{84.7}  & \textbf{72.8} \\
    \hline
\end{tabular}%
}
\end{center}
\caption{Classification accuracies (\%) on Office-Home for ResNet50-based methods.}
\label{tab:tab2}
\end{table}%

As shown in Table \ref{tab:tab2}, our method achieves the latest performance (72.8\%) and is higher than the second best NRC by a margin of 0.6\% on Office-Home, achieving the best/second-best results on 10 out of 12 individual tasks. Our method is even superior to some of the traditional domain adaptation methods which require source data. This can be attributed to the fact that due to the increased amount of data, more finely polycentric clustering and more comprehensive structure information is available to support our approach.

\begin{table}
\begin{center}
\setlength{\tabcolsep}{2pt}
\renewcommand\arraystretch{0.9}
\resizebox{\linewidth}{!}{
    \begin{tabular}{lcccccccccccccc}
    \hline
    \multicolumn{1}{c}{Method} & 
    SF & 
    \multicolumn{1}{c}{plane} & 
    \multicolumn{1}{c}{bicycle} & 
    \multicolumn{1}{c}{bus} & 
    \multicolumn{1}{c}{car} & 
    \multicolumn{1}{c}{horse} & 
    \multicolumn{1}{c}{knife} & 
    \multicolumn{1}{c}{motor} &
    \multicolumn{1}{c}{person} & 
    \multicolumn{1}{c}{plant} & 
    \multicolumn{1}{c}{sktbrd} & 
    \multicolumn{1}{c}{train} & 
    \multicolumn{1}{c}{truck} & 
    \multicolumn{1}{c}{Per-class} \\
    \hline
    ResNet-101(2016)\cite{he2016deep} & \ding{55} & 55.1  & 53.3  & 61.9  & 59.1  & 80.6  & 17.9  & 79.7  & 31.2  & 81.0  & 26.5  & 73.5  & 8.5  & 52.4  \\
    CDAN(2018)\cite{long2018conditional} & \ding{55} & 85.2  & 66.9  & 83.0  & 50.8  & 84.2  & 74.9  & 88.1  & 74.5  & 83.4  & 76.0  & 81.9  & 38.0  & 73.9  \\
    SAFN(2019)\cite{xu2019larger} & \ding{55} & 93.6  & 61.3  & 84.1  & 70.6  & 94.1  & 79.0  & 91.8  & 79.6  & 89.9  & 55.6  & 89.0  & 24.4  & 76.1  \\
    SWD(2019)\cite{lee2019sliced} & \ding{55} & 90.8  & 82.5  & 81.7  & 70.5  & 91.7  & 69.5  & 86.3  & 77.5  & 87.4  & 63.6  & 85.6  & 29.2  & 76.4  \\
    MCC(2020)\cite{jin2020minimum} & \ding{55} & 88.7  & 80.3  & 80.5  & 71.5  & 90.1  & 93.2  & 85.0  & 71.6  & 89.4  & 73.8  & 85.0  & 36.9  & 78.8  \\
    RWOT(2020)\cite{xu2020reliable} & \ding{55} & 95.1  & 80.3  & 83.7  & 90.0  & 92.4  & 68.0  & 92.5  & 82.2  & 87.9  & 78.4  & 90.4  & 68.2  & 84.1  \\
    \hline
    Source-only & \checkmark & 64.1  & 24.9  & 53.0  & 66.5  & 67.9  & 9.1  & 84.5  & 21.1  & 62.8  & 29.8  & 83.5  & 9.3  & 48.0  \\
    SFDA(2021)\cite{kim2021domain} & \checkmark & 86.9  & 81.7  & \uline{84.6}  & 63.9  & 93.1  & 91.4  & 86.6  & 71.9  & 84.5  & 58.2  & 74.5  & 42.7  & 76.7  \\
    SHOT(2020a)\cite{liang2020we} & \checkmark & 94.3  & \uline{88.5}  & 80.1  & 57.3  & 93.1  & 94.9  & 80.7  & 80.3  & 91.5  & 89.1  & 86.3  & 58.2  & 82.9  \\
    3C-GAN(2020)\cite{li2020model} & \checkmark & 94.8  & 73.4  & 68.8  & \textbf{74.8} & 93.1  & 95.4  & \uline{88.6}  & \textbf{84.7} & 89.1  & 84.7  & 83.5  & 48.1  & 81.6  \\
    G-SFDA(2021b)\cite{yang2021generalized} & \checkmark & \uline{96.1}  & 88.3  & \textbf{85.5} & \uline{74.1}  & 97.1  & 95.4  & \textbf{89.5} & 79.4  & \textbf{95.4} & \textbf{92.9} & \uline{89.1}  & 42.6  & \uline{85.5}  \\
    HCL(2021a)\cite{tang2021nearest} & \checkmark & 93.3  & 85.4  & 80.7  & 68.5  & 91.0  & 88.1  & 86.0  & 78.6  & 86.6  & 88.8  & 80.0  & \textbf{74.7} & 83.5  \\
    \bf{ours} & \checkmark & \textbf{96.2} & \textbf{89.8} & 82.5  & 61.0  & \uline{95.3}  & \textbf{96.4} & 87.5 & \uline{81.8}  & 91.7  & \uline{92.3}  & 86.2  & \uline{66.4} & \textbf{85.6} \\
    \hline
    \end{tabular}
    }
\end{center}
\setlength{\belowcaptionskip}{-0.4cm} 
\caption{Classification accuracies (\%) on VisDA-C for ResNet101-based methods.}
\label{tab:tab3}
\end{table}

To further demonstrate the effectiveness of our proposed PCSR, we conduct evaluation experiments on the large dataset VisDA-C and illustrate the results in Table \ref{tab:tab3}. Our method significantly outperforms SHOT, surpassing it by 2.7\%. We can find that class-balanced performance has been improved with our strategy. Especially for the challenging class ‘truck’, our method achieves 66.4\%, which outperforms SHOT applied monocentric clustering by 23.7\%. The reason is that the polycentric clustering strategy introduces more fine-grained feature clustering centers and the generalization ability of the target model is improved by structural regularization. The results demonstrate the effectiveness of our approach, and our method also outperforms domain adaptation methods with access to source data on both Office-Home and VisDA-C.

\subsection{Ablation studies}
{\bf Number of clustering centers P}. In Figure \ref{fig:figP}, we show the results using different P$\in$\{1,2,3,4,
5,6\} on three datasets. When P is set to 1, it is a generalized single-centroid clustering strategy. It can be discovered that when P is set to 3, accuracy can go up to the best result on Office-31 and Office-Home, whereas the best result for VisDA-C is obtained when P is set to 4. The data size of VisDA-C is larger than Office-31 and Office-Home, and the adaptation is performed from the synthetic images to the real images on VisDA-C, there is a great discrepancy in the feature distribution between them. Therefore, more clustering centers should be set to achieve better performance. It can be seen that the polycentric clustering strategy introduce more fine-grained feature clustering centers for each class, which allows the model to assign more accurate pseudo-labels for those hard transfer data. This demonstrates that it is necessary to implement intra-class polycentric clustering.

{\bf Ablation study of hyper-parameter $\beta$}. In Eq.(\ref{eq8}), $\beta$ is an empirical hyper-parameter, and we conduct an ablation experiment which is shown in Fig. \ref{fig:fig3}. Some different $\beta$ in [0,5, 1.5] are set for the experiment, and it can be seen that the value of $\beta$ is insensitive to any change, so we select $\beta$ = 1 in the paper.

\makeatletter\def\@captype{figure}\makeatother
\vspace{0.2cm}
\begin{minipage}{0.45\linewidth}
    \centering
    \setlength{\abovecaptionskip}{-0.8cm}
    \includegraphics[width=\linewidth]{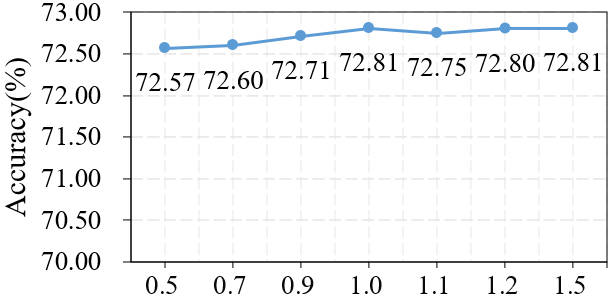}
    \caption{The ablation study of $\beta$}
    \label{fig:fig3}
    \vspace{0.2cm}
\end{minipage}
\hfill
\makeatletter\def\@captype{table}\makeatother
\begin{minipage}{0.45\linewidth}
    \centering
    \vspace{0.2cm}
    \begin{tabular}{ccc|c}
    \hline
    $\mathcal{L}_{im}$ & $\mathcal{L}_{pcc}$ & $\mathcal{L}_{mix}$ & Avg.  \\
    \hline
    & & & 59.6  \\
    \checkmark & & & 70.5  \\
    \checkmark & \checkmark & & 72.1  \\
    \checkmark &  & \checkmark & 72.5  \\
    \checkmark & \checkmark & \checkmark & 72.8  \\
    \hline
    \end{tabular}
    \setlength{\abovecaptionskip}{0.2cm}
    \caption{Ablation of the losses on Office-Home.}
    \label{tab:tab4}
    \vspace{0.2cm}
\end{minipage}

{\bf Ablation study on losses}. We validate the effectiveness of our methods on Office-Home. Results are shown in Table \ref{tab:tab4}. The classification accuracy is 59.6\% when the source-only model is used. We start with applying the information maximization loss, where makes the classification output of the target features becomes more certainty and more global diversity. This achieves 70.5\% accuracy. In the third row, based on the information maximization loss, with the intra-class polycentric feature clustering, more accurate pseudo labels are obtained, and the performance increases by 12.5\% to 72.1\%. And using structural regularization, the model is enforced  to maintain consistency, and the performance achieves 72.5\%. The model’s performance is optimized when all three are used simultaneously. This demonstrates the validity of each loss function.

{\bf t-SNE visualization}. To visually the effectiveness of our method, we compare the t-SNEs embedding the features extracted by ResNet-50, SHOT and our method on Ar$\rightarrow${Cl} in Office-Home. As shown in Figure \ref{fig:figtSNE}, where the source and target domain features are expected to cluster independently. We observe that the features in the target domain become more structured after adaptation, and the source and target domains are better aligned via our method. This result clearly demonstrates that it is possible to reduce the discrepancy between two different domains even without accessing source data.
\begin{figure}
\centering
\setlength{\abovecaptionskip}{0.cm}
\begin{tabular}{ccc}    
\includegraphics[height=3.3cm, width=3.8cm]{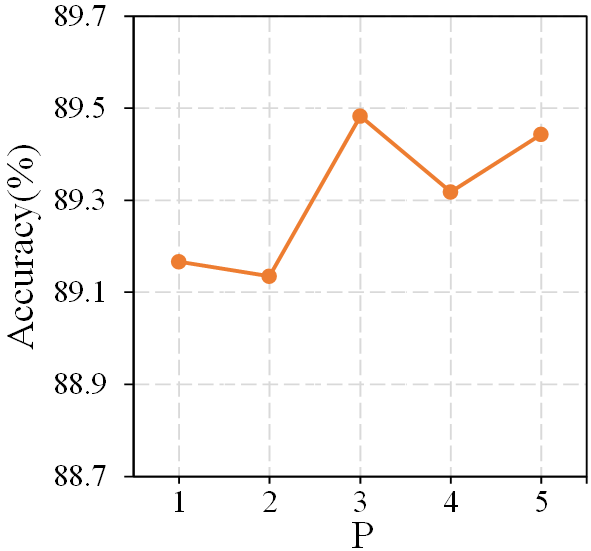} &
\includegraphics[height=3.3cm,width=3.8cm]{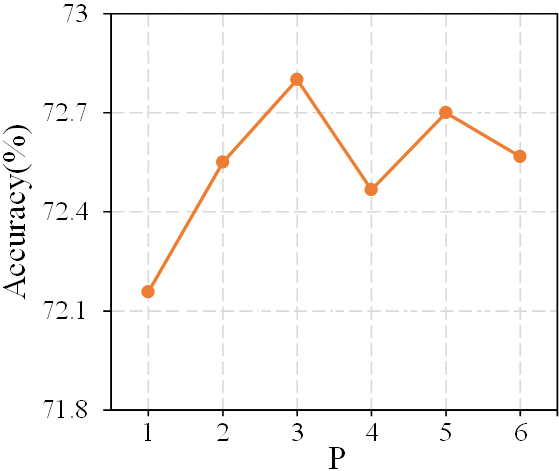}&
\includegraphics[height=3.3cm,width=3.8cm]{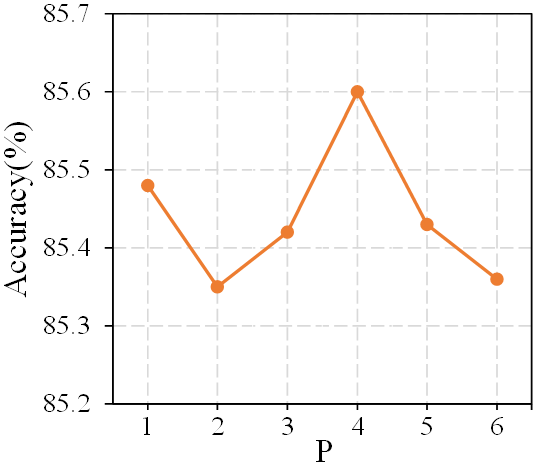}\\
 \qquad{(a)}&\qquad{(b)}&\qquad{(c)}
\end{tabular}
\caption{(a), (b) and (c) presents the accuracy on Office-31, Office-Home and VisDA-C when P takes different values, respectively.}
\label{fig:figP}
\end{figure}

\begin{figure}
\centering
\begin{tabular}{ccc}
\bmvaHangBox{\includegraphics[width=4cm]{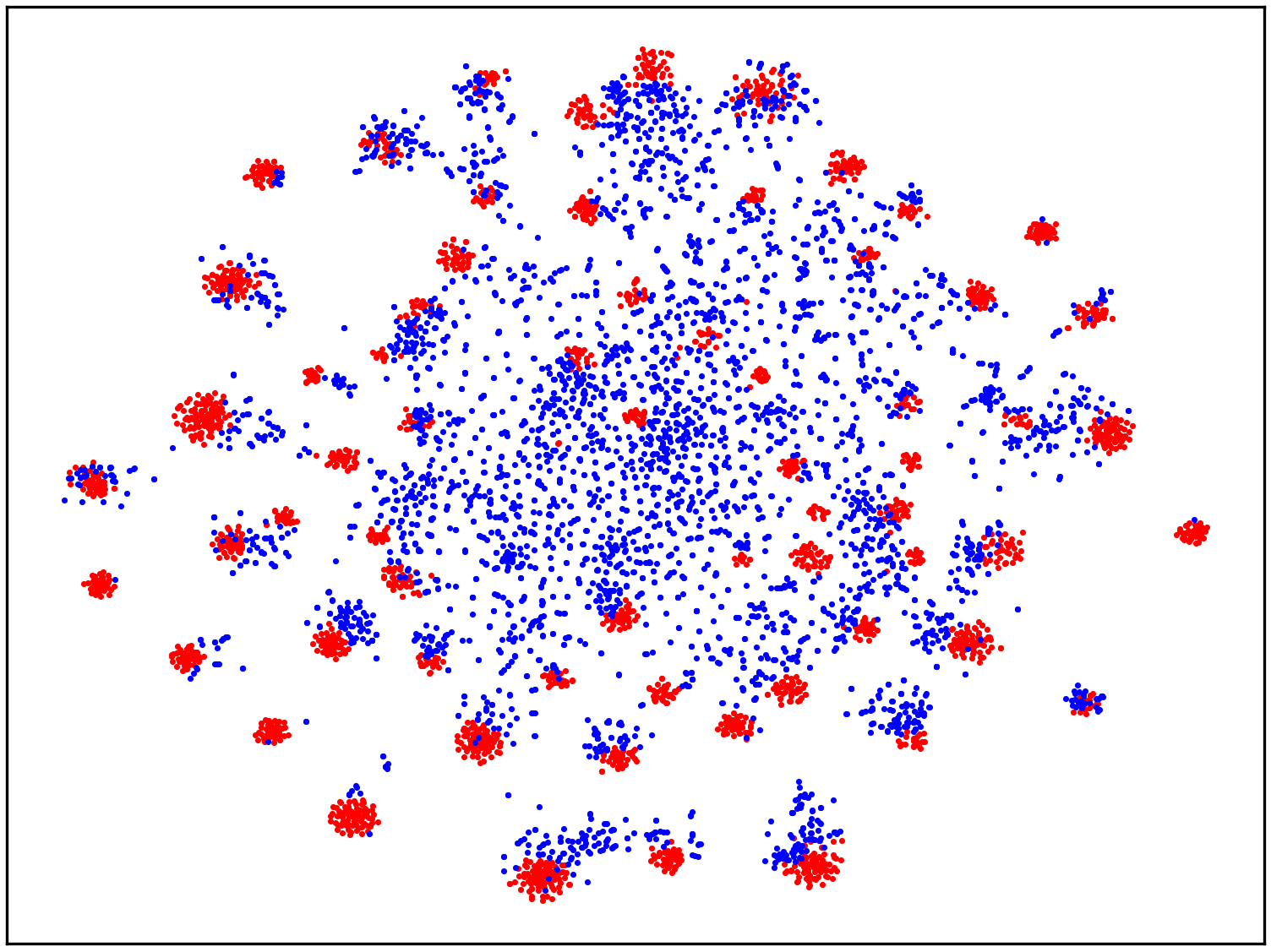}} &
\bmvaHangBox{\includegraphics[width=4cm]{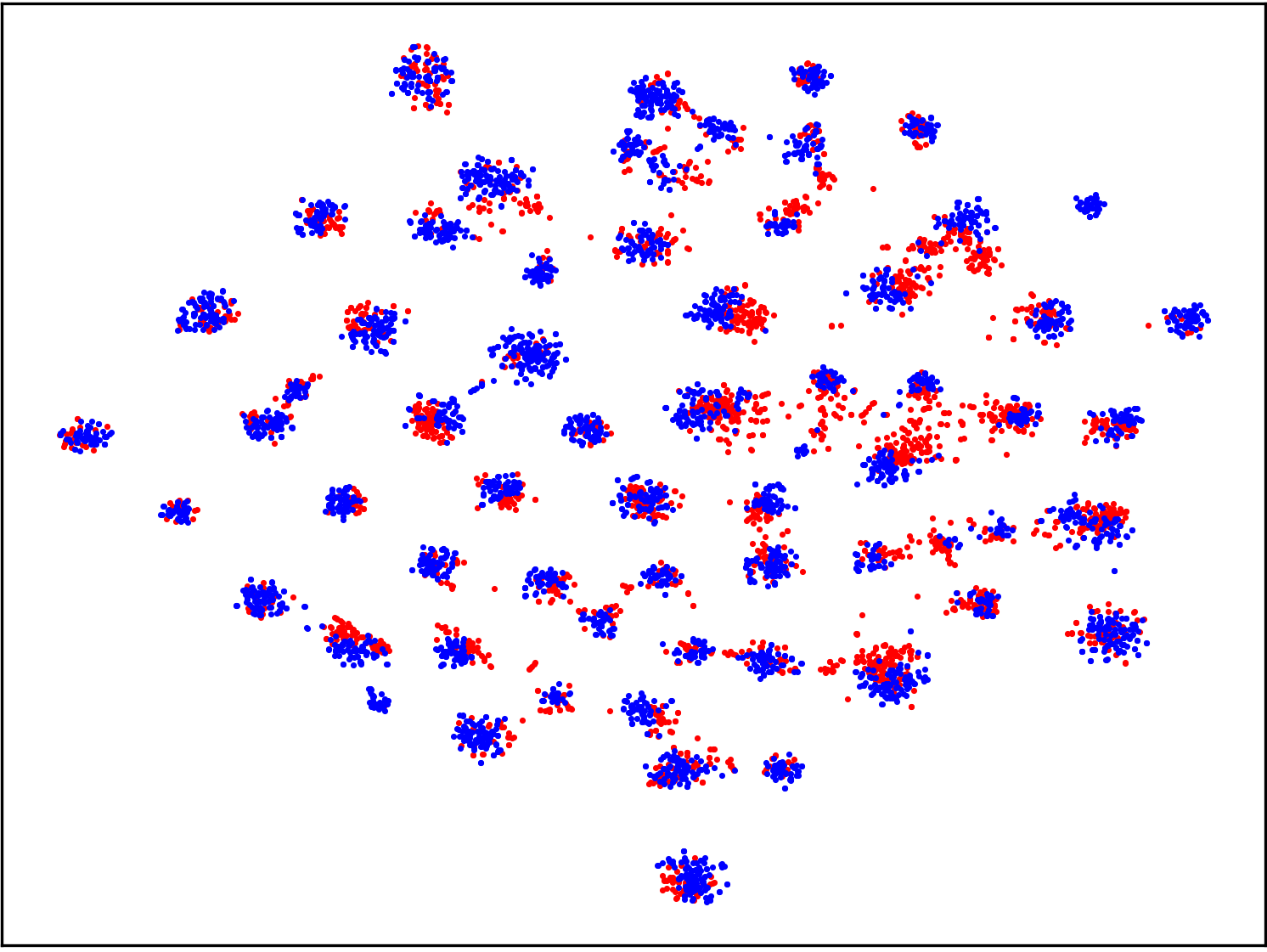}} &
\bmvaHangBox{\includegraphics[width=4cm]{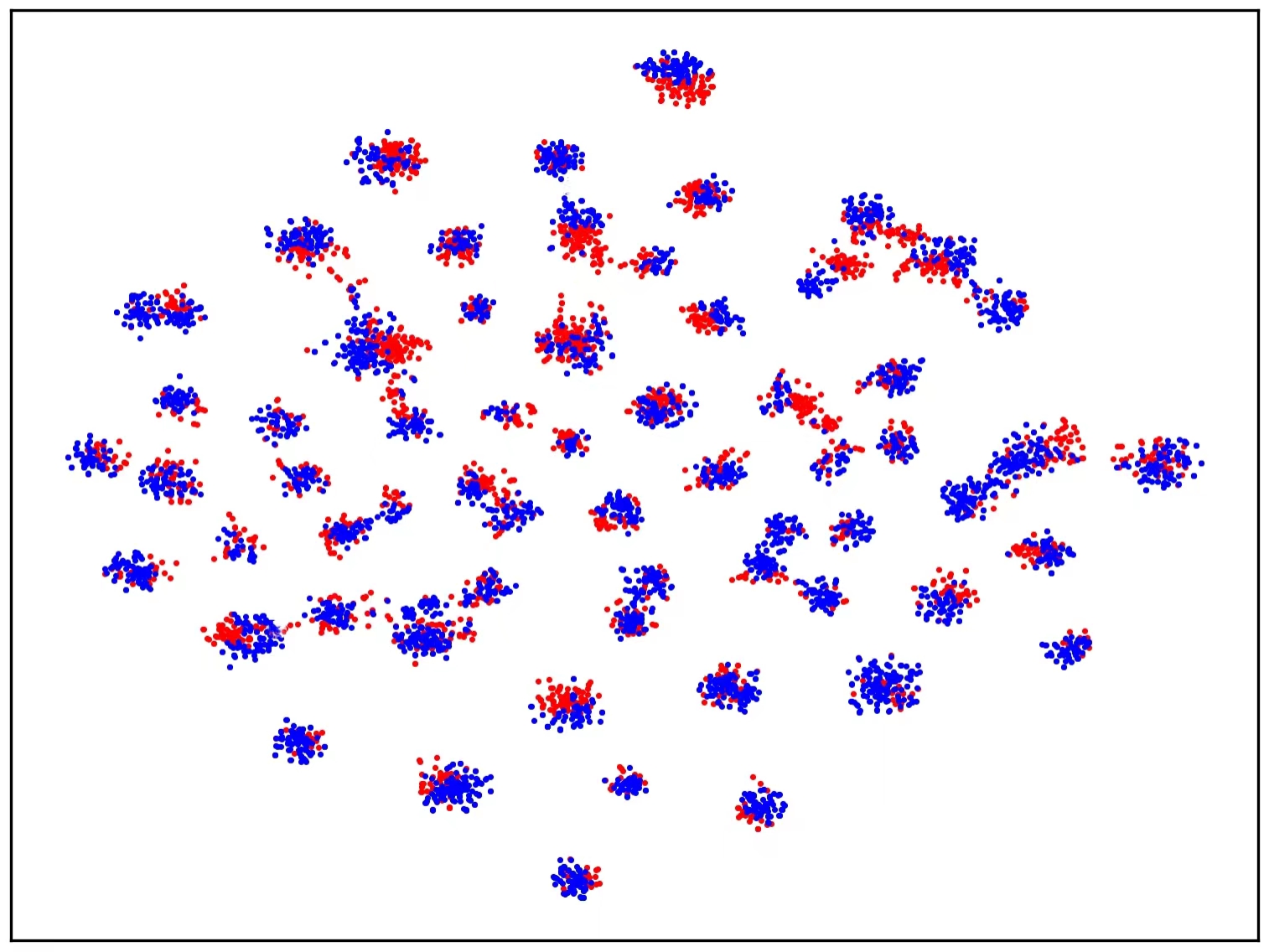}}\\
(a)&(b)&(c)
\end{tabular}
\caption{The t-SNE feature visualizations for task Ar→Cl on Office-Home. (a), (b) and (c) presents alignment between the source data and the target data by source-only model, SHOT and our PCSR method, respectively. Red/blue denote the source/target domains.}
\label{fig:figtSNE}
\end{figure}
\vspace{0.2cm}

\section{Conclusion}
In this paper, we have proposed a polycentric clustering and structure regularization (PCSR) strategy for source-free domain adaptation. Specifically different from the previous monocentric clustering, our PCSR strategy reduced the negative transfer of hard data in the target domain by considering intra-class polycentric clustering through inter-class-balanced sampling. In addition, structural regularization of the target domain interpolates the target data for consistent training, and improves the model’s robustness. The experimental results on three benchmark datasets have demonstrated the effectiveness of our approach. For future work, we intend to apply the method to other vision tasks, such as semantic segmentation and target detection.

\normalem
\bibliography{egbib}

\begin{thebibliography}{52}
\providecommand{\natexlab}[1]{#1}
\providecommand{\url}[1]{\texttt{#1}}
\expandafter\ifx\csname urlstyle\endcsname\relax
  \providecommand{\doi}[1]{doi: #1}\else
  \providecommand{\doi}{doi: \begingroup \urlstyle{rm}\Url}\fi

\bibitem[Arazo et~al.(2020)Arazo, Ortego, Albert, O’Connor, and
  McGuinness]{arazo2020pseudo}
Eric Arazo, Diego Ortego, Paul Albert, Noel~E O’Connor, and Kevin McGuinness.
\newblock Pseudo-labeling and confirmation bias in deep semi-supervised
  learning.
\newblock In \emph{2020 International Joint Conference on Neural Networks
  (IJCNN)}, pages 1--8. IEEE, 2020.

\bibitem[Bousmalis et~al.(2016)Bousmalis, Trigeorgis, Silberman, Krishnan, and
  Erhan]{bousmalis2016domain}
Konstantinos Bousmalis, George Trigeorgis, Nathan Silberman, Dilip Krishnan,
  and Dumitru Erhan.
\newblock Domain separation networks.
\newblock \emph{Advances in neural information processing systems}, 29, 2016.

\bibitem[Chen et~al.(2018)Chen, Li, Sakaridis, Dai, and
  Van~Gool]{chen2018domain}
Yuhua Chen, Wen Li, Christos Sakaridis, Dengxin Dai, and Luc Van~Gool.
\newblock Domain adaptive faster r-cnn for object detection in the wild.
\newblock In \emph{Proceedings of the IEEE conference on computer vision and
  pattern recognition}, pages 3339--3348, 2018.

\bibitem[Cui et~al.(2020)Cui, Wang, Zhuo, Li, Huang, and Tian]{cui2020towards}
Shuhao Cui, Shuhui Wang, Junbao Zhuo, Liang Li, Qingming Huang, and Qi~Tian.
\newblock Towards discriminability and diversity: Batch nuclear-norm
  maximization under label insufficient situations.
\newblock In \emph{Proceedings of the IEEE/CVF Conference on Computer Vision
  and Pattern Recognition}, pages 3941--3950, 2020.

\bibitem[Cui et~al.(2021)Cui, Wang, Zhuo, Li, Huang, and Tian]{cui2021fast}
Shuhao Cui, Shuhui Wang, Junbao Zhuo, Liang Li, Qingming Huang, and Qi~Tian.
\newblock Fast batch nuclear-norm maximization and minimization for robust
  domain adaptation.
\newblock \emph{arXiv preprint arXiv:2107.06154}, 2021.

\bibitem[Deng et~al.(2009)Deng, Dong, Socher, Li, Li, and
  Fei-Fei]{deng2009imagenet}
Jia Deng, Wei Dong, Richard Socher, Li-Jia Li, Kai Li, and Li~Fei-Fei.
\newblock Imagenet: A large-scale hierarchical image database.
\newblock In \emph{2009 IEEE conference on computer vision and pattern
  recognition}, pages 248--255. Ieee, 2009.

\bibitem[Deng et~al.(2018)Deng, Zheng, Ye, Kang, Yang, and Jiao]{deng2018image}
Weijian Deng, Liang Zheng, Qixiang Ye, Guoliang Kang, Yi~Yang, and Jianbin
  Jiao.
\newblock Image-image domain adaptation with preserved self-similarity and
  domain-dissimilarity for person re-identification.
\newblock In \emph{Proceedings of the IEEE conference on computer vision and
  pattern recognition}, pages 994--1003, 2018.

\bibitem[Deng et~al.(2019)Deng, Luo, and Zhu]{deng2019cluster}
Zhijie Deng, Yucen Luo, and Jun Zhu.
\newblock Cluster alignment with a teacher for unsupervised domain adaptation.
\newblock In \emph{Proceedings of the IEEE/CVF international conference on
  computer vision}, pages 9944--9953, 2019.

\bibitem[Ganin and Lempitsky(2015)]{ganin2015unsupervised}
Yaroslav Ganin and Victor Lempitsky.
\newblock Unsupervised domain adaptation by backpropagation.
\newblock In \emph{International conference on machine learning}, pages
  1180--1189. PMLR, 2015.

\bibitem[Ganin et~al.(2016)Ganin, Ustinova, Ajakan, Germain, Larochelle,
  Laviolette, Marchand, and Lempitsky]{ganin2016domain}
Yaroslav Ganin, Evgeniya Ustinova, Hana Ajakan, Pascal Germain, Hugo
  Larochelle, Fran{\c{c}}ois Laviolette, Mario Marchand, and Victor Lempitsky.
\newblock Domain-adversarial training of neural networks.
\newblock \emph{The journal of machine learning research}, 17\penalty0
  (1):\penalty0 2096--2030, 2016.

\bibitem[He et~al.(2016)He, Zhang, Ren, and Sun]{he2016deep}
Kaiming He, Xiangyu Zhang, Shaoqing Ren, and Jian Sun.
\newblock Deep residual learning for image recognition.
\newblock In \emph{Proceedings of the IEEE conference on computer vision and
  pattern recognition}, pages 770--778, 2016.

\bibitem[Hu et~al.(2017)Hu, Miyato, Tokui, Matsumoto, and
  Sugiyama]{hu2017learning}
Weihua Hu, Takeru Miyato, Seiya Tokui, Eiichi Matsumoto, and Masashi Sugiyama.
\newblock Learning discrete representations via information maximizing
  self-augmented training.
\newblock In \emph{International conference on machine learning}, pages
  1558--1567. PMLR, 2017.

\bibitem[Jin et~al.(2020)Jin, Wang, Long, and Wang]{jin2020minimum}
Ying Jin, Ximei Wang, Mingsheng Long, and Jianmin Wang.
\newblock Minimum class confusion for versatile domain adaptation.
\newblock In \emph{European Conference on Computer Vision}, pages 464--480.
  Springer, 2020.

\bibitem[Kim et~al.(2021)Kim, Cho, Han, Panda, and Hong]{kim2021domain}
Youngeun Kim, Donghyeon Cho, Kyeongtak Han, Priyadarshini Panda, and Sungeun
  Hong.
\newblock Domain adaptation without source data.
\newblock \emph{IEEE Transactions on Artificial Intelligence}, 2\penalty0
  (6):\penalty0 508--518, 2021.

\bibitem[Kundu et~al.(2020)Kundu, Venkat, Babu, et~al.]{kundu2020universal}
Jogendra~Nath Kundu, Naveen Venkat, R~Venkatesh Babu, et~al.
\newblock Universal source-free domain adaptation.
\newblock In \emph{Proceedings of the IEEE/CVF Conference on Computer Vision
  and Pattern Recognition}, pages 4544--4553, 2020.

\bibitem[Le et~al.(2021)Le, Nguyen, Ho, Bui, and Phung]{le2021lamda}
Trung Le, Tuan Nguyen, Nhat Ho, Hung Bui, and Dinh Phung.
\newblock Lamda: Label matching deep domain adaptation.
\newblock In \emph{International Conference on Machine Learning}, pages
  6043--6054. PMLR, 2021.

\bibitem[Lee et~al.(2019)Lee, Batra, Baig, and Ulbricht]{lee2019sliced}
Chen-Yu Lee, Tanmay Batra, Mohammad~Haris Baig, and Daniel Ulbricht.
\newblock Sliced wasserstein discrepancy for unsupervised domain adaptation.
\newblock In \emph{Proceedings of the IEEE/CVF Conference on Computer Vision
  and Pattern Recognition}, pages 10285--10295, 2019.

\bibitem[Li et~al.(2020)Li, Jiao, Cao, Wong, and Wu]{li2020model}
Rui Li, Qianfen Jiao, Wenming Cao, Hau-San Wong, and Si~Wu.
\newblock Model adaptation: Unsupervised domain adaptation without source data.
\newblock In \emph{Proceedings of the IEEE/CVF Conference on Computer Vision
  and Pattern Recognition}, pages 9641--9650, 2020.

\bibitem[Liang et~al.(2020)Liang, Hu, and Feng]{liang2020we}
Jian Liang, Dapeng Hu, and Jiashi Feng.
\newblock Do we really need to access the source data? source hypothesis
  transfer for unsupervised domain adaptation.
\newblock In \emph{International Conference on Machine Learning}, pages
  6028--6039. PMLR, 2020.

\bibitem[Liu et~al.(2021)Liu, Zhang, and Wang]{liu2021source}
Yuang Liu, Wei Zhang, and Jun Wang.
\newblock Source-free domain adaptation for semantic segmentation.
\newblock In \emph{Proceedings of the IEEE/CVF Conference on Computer Vision
  and Pattern Recognition}, pages 1215--1224, 2021.

\bibitem[Long et~al.(2015)Long, Cao, Wang, and Jordan]{long2015learning}
Mingsheng Long, Yue Cao, Jianmin Wang, and Michael Jordan.
\newblock Learning transferable features with deep adaptation networks.
\newblock In \emph{International conference on machine learning}, pages
  97--105. PMLR, 2015.

\bibitem[Long et~al.(2016)Long, Zhu, Wang, and Jordan]{long2016unsupervised}
Mingsheng Long, Han Zhu, Jianmin Wang, and Michael~I Jordan.
\newblock Unsupervised domain adaptation with residual transfer networks.
\newblock \emph{Advances in neural information processing systems}, 29, 2016.

\bibitem[Long et~al.(2017)Long, Zhu, Wang, and Jordan]{long2017deep}
Mingsheng Long, Han Zhu, Jianmin Wang, and Michael~I Jordan.
\newblock Deep transfer learning with joint adaptation networks.
\newblock In \emph{International conference on machine learning}, pages
  2208--2217. PMLR, 2017.

\bibitem[Long et~al.(2018{\natexlab{a}})Long, Cao, Cao, Wang, and
  Jordan]{long2018transferable}
Mingsheng Long, Yue Cao, Zhangjie Cao, Jianmin Wang, and Michael~I Jordan.
\newblock Transferable representation learning with deep adaptation networks.
\newblock \emph{IEEE transactions on pattern analysis and machine
  intelligence}, 41\penalty0 (12):\penalty0 3071--3085, 2018{\natexlab{a}}.

\bibitem[Long et~al.(2018{\natexlab{b}})Long, Cao, Wang, and
  Jordan]{long2018conditional}
Mingsheng Long, Zhangjie Cao, Jianmin Wang, and Michael~I Jordan.
\newblock Conditional adversarial domain adaptation.
\newblock \emph{Advances in neural information processing systems}, 31,
  2018{\natexlab{b}}.

\bibitem[MacQueen(1967)]{macqueen1967classification}
J~MacQueen.
\newblock Classification and analysis of multivariate observations.
\newblock In \emph{5th Berkeley Symp. Math. Statist. Probability}, pages
  281--297, 1967.

\bibitem[Murez et~al.(2018)Murez, Kolouri, Kriegman, Ramamoorthi, and
  Kim]{murez2018image}
Zak Murez, Soheil Kolouri, David Kriegman, Ravi Ramamoorthi, and Kyungnam Kim.
\newblock Image to image translation for domain adaptation.
\newblock In \emph{Proceedings of the IEEE Conference on Computer Vision and
  Pattern Recognition}, pages 4500--4509, 2018.

\bibitem[Pan and Yang(2009)]{pan2009survey}
Sinno~Jialin Pan and Qiang Yang.
\newblock A survey on transfer learning.
\newblock \emph{IEEE Transactions on knowledge and data engineering},
  22\penalty0 (10):\penalty0 1345--1359, 2009.

\bibitem[Peng et~al.(2017)Peng, Usman, Kaushik, Hoffman, Wang, and
  Saenko]{peng2017visda}
Xingchao Peng, Ben Usman, Neela Kaushik, Judy Hoffman, Dequan Wang, and Kate
  Saenko.
\newblock Visda: The visual domain adaptation challenge.
\newblock \emph{arXiv preprint arXiv:1710.06924}, 2017.

\bibitem[Peng et~al.(2019)Peng, Bai, Xia, Huang, Saenko, and
  Wang]{peng2019moment}
Xingchao Peng, Qinxun Bai, Xide Xia, Zijun Huang, Kate Saenko, and Bo~Wang.
\newblock Moment matching for multi-source domain adaptation.
\newblock In \emph{Proceedings of the IEEE/CVF international conference on
  computer vision}, pages 1406--1415, 2019.

\bibitem[Qiu et~al.(2021)Qiu, Zhang, Lin, Niu, and Tan]{2021Source}
Z.~Qiu, Y.~Zhang, H.~Lin, S.~Niu, and M.~Tan.
\newblock Source-free domain adaptation via avatar prototype generation and
  adaptation.
\newblock 2021.

\bibitem[Saenko et~al.(2010)Saenko, Kulis, Fritz, and
  Darrell]{saenko2010adapting}
Kate Saenko, Brian Kulis, Mario Fritz, and Trevor Darrell.
\newblock Adapting visual category models to new domains.
\newblock In \emph{European conference on computer vision}, pages 213--226.
  Springer, 2010.

\bibitem[Sahoo et~al.(2020)Sahoo, Shanmugam, and Guttag]{sahoo2020unsupervised}
Roshni Sahoo, Divya Shanmugam, and John Guttag.
\newblock Unsupervised domain adaptation in the absence of source data.
\newblock \emph{arXiv preprint arXiv:2007.10233}, 2020.

\bibitem[Shu et~al.(2018)Shu, Bui, Narui, and Ermon]{shu2018dirt}
Rui Shu, Hung~H Bui, Hirokazu Narui, and Stefano Ermon.
\newblock A dirt-t approach to unsupervised domain adaptation.
\newblock \emph{arXiv preprint arXiv:1802.08735}, 2018.

\bibitem[Su et~al.(2020)Su, Wang, Zeng, Tang, Chen, Qiu, and
  Wang]{su2020adapting}
Peng Su, Kun Wang, Xingyu Zeng, Shixiang Tang, Dapeng Chen, Di~Qiu, and
  Xiaogang Wang.
\newblock Adapting object detectors with conditional domain normalization.
\newblock In \emph{European Conference on Computer Vision}, pages 403--419.
  Springer, 2020.

\bibitem[Tang et~al.(2020)Tang, Chen, and Jia]{tang2020unsupervised}
Hui Tang, Ke~Chen, and Kui Jia.
\newblock Unsupervised domain adaptation via structurally regularized deep
  clustering.
\newblock In \emph{Proceedings of the IEEE/CVF conference on computer vision
  and pattern recognition}, pages 8725--8735, 2020.

\bibitem[Tang et~al.(2021)Tang, Yang, Ma, Hendrich, Zeng, Ge, Zhang, and
  Zhang]{tang2021nearest}
Song Tang, Yan Yang, Zhiyuan Ma, Norman Hendrich, Fanyu Zeng, Shuzhi~Sam Ge,
  Changshui Zhang, and Jianwei Zhang.
\newblock Nearest neighborhood-based deep clustering for source data-absent
  unsupervised domain adaptation.
\newblock \emph{arXiv preprint arXiv:2107.12585}, 2021.

\bibitem[Tian et~al.(2021)Tian, Zhang, Li, and Xu]{tian2021vdm}
Jiayi Tian, Jing Zhang, Wen Li, and Dong Xu.
\newblock Vdm-da: Virtual domain modeling for source data-free domain
  adaptation.
\newblock \emph{IEEE Transactions on Circuits and Systems for Video
  Technology}, 2021.

\bibitem[Toldo et~al.(2020)Toldo, Maracani, Michieli, and
  Zanuttigh]{toldo2020unsupervised}
Marco Toldo, Andrea Maracani, Umberto Michieli, and Pietro Zanuttigh.
\newblock Unsupervised domain adaptation in semantic segmentation: a review.
\newblock \emph{Technologies}, 8\penalty0 (2):\penalty0 35, 2020.

\bibitem[Venkateswara et~al.(2017)Venkateswara, Eusebio, Chakraborty, and
  Panchanathan]{venkateswara2017deep}
Hemanth Venkateswara, Jose Eusebio, Shayok Chakraborty, and Sethuraman
  Panchanathan.
\newblock Deep hashing network for unsupervised domain adaptation.
\newblock In \emph{Proceedings of the IEEE conference on computer vision and
  pattern recognition}, pages 5018--5027, 2017.

\bibitem[Verma et~al.(2019)Verma, Kawaguchi, Lamb, Kannala, Bengio, and
  Lopez-Paz]{verma2019interpolation}
Vikas Verma, Kenji Kawaguchi, Alex Lamb, Juho Kannala, Yoshua Bengio, and David
  Lopez-Paz.
\newblock Interpolation consistency training for semi-supervised learning.
\newblock \emph{arXiv preprint arXiv:1903.03825}, 2019.

\bibitem[Wang et~al.(2019)Wang, Wang, Zheng, Wu, Zeng, and
  Satoh]{wang2019beyond}
Zheng Wang, Zhixiang Wang, Yinqiang Zheng, Yang Wu, Wenjun Zeng, and Shin'ichi
  Satoh.
\newblock Beyond intra-modality: A survey of heterogeneous person
  re-identification.
\newblock \emph{arXiv preprint arXiv:1905.10048}, 2019.

\bibitem[Wu et~al.(2020)Wu, Inkpen, and El-Roby]{wu2020dual}
Yuan Wu, Diana Inkpen, and Ahmed El-Roby.
\newblock Dual mixup regularized learning for adversarial domain adaptation.
\newblock In \emph{European Conference on Computer Vision}, pages 540--555.
  Springer, 2020.

\bibitem[Xia et~al.(2021)Xia, Zhao, and Ding]{xia2021adaptive}
Haifeng Xia, Handong Zhao, and Zhengming Ding.
\newblock Adaptive adversarial network for source-free domain adaptation.
\newblock In \emph{Proceedings of the IEEE/CVF International Conference on
  Computer Vision}, pages 9010--9019, 2021.

\bibitem[Xu et~al.(2020)Xu, Liu, Wang, Chen, and Wang]{xu2020reliable}
Renjun Xu, Pelen Liu, Liyan Wang, Chao Chen, and Jindong Wang.
\newblock Reliable weighted optimal transport for unsupervised domain
  adaptation.
\newblock In \emph{Proceedings of the IEEE/CVF conference on computer vision
  and pattern recognition}, pages 4394--4403, 2020.

\bibitem[Xu et~al.(2019)Xu, Li, Yang, and Lin]{xu2019larger}
Ruijia Xu, Guanbin Li, Jihan Yang, and Liang Lin.
\newblock Larger norm more transferable: An adaptive feature norm approach for
  unsupervised domain adaptation.
\newblock In \emph{Proceedings of the IEEE/CVF International Conference on
  Computer Vision}, pages 1426--1435, 2019.

\bibitem[Yang et~al.(2020)Yang, Wang, van~de Weijer, Herranz, and
  Jui]{yang2020unsupervised}
Shiqi Yang, Yaxing Wang, Joost van~de Weijer, Luis Herranz, and Shangling Jui.
\newblock Unsupervised domain adaptation without source data by casting a bait.
\newblock \emph{arXiv preprint arXiv:2010.12427}, 1\penalty0 (2):\penalty0 5,
  2020.

\bibitem[Yang et~al.(2021{\natexlab{a}})Yang, van~de Weijer, Herranz, Jui,
  et~al.]{yang2021exploiting}
Shiqi Yang, Joost van~de Weijer, Luis Herranz, Shangling Jui, et~al.
\newblock Exploiting the intrinsic neighborhood structure for source-free
  domain adaptation.
\newblock \emph{Advances in Neural Information Processing Systems},
  34:\penalty0 29393--29405, 2021{\natexlab{a}}.

\bibitem[Yang et~al.(2021{\natexlab{b}})Yang, Wang, van~de Weijer, Herranz, and
  Jui]{yang2021generalized}
Shiqi Yang, Yaxing Wang, Joost van~de Weijer, Luis Herranz, and Shangling Jui.
\newblock Generalized source-free domain adaptation.
\newblock In \emph{Proceedings of the IEEE/CVF International Conference on
  Computer Vision}, pages 8978--8987, 2021{\natexlab{b}}.

\bibitem[Zhang et~al.(2017)Zhang, Cisse, Dauphin, and
  Lopez-Paz]{zhang2017mixup}
Hongyi Zhang, Moustapha Cisse, Yann~N Dauphin, and David Lopez-Paz.
\newblock mixup: Beyond empirical risk minimization.
\newblock \emph{arXiv preprint arXiv:1710.09412}, 2017.

\bibitem[Zhang et~al.(2019)Zhang, Tang, Jia, and Tan]{zhang2019domain}
Yabin Zhang, Hui Tang, Kui Jia, and Mingkui Tan.
\newblock Domain-symmetric networks for adversarial domain adaptation.
\newblock In \emph{Proceedings of the IEEE/CVF conference on computer vision
  and pattern recognition}, pages 5031--5040, 2019.

\bibitem[Zou et~al.(2018)Zou, Yu, Kumar, and Wang]{zou2018unsupervised}
Yang Zou, Zhiding Yu, BVK Kumar, and Jinsong Wang.
\newblock Unsupervised domain adaptation for semantic segmentation via
  class-balanced self-training.
\newblock In \emph{Proceedings of the European conference on computer vision
  (ECCV)}, pages 289--305, 2018.

\end{thebibliography}
\end{document}